\documentclass[letterpaper, 10 pt, conference]{ieeeconf}
\usepackage{hyperref}
\usepackage{diagbox}
\usepackage{graphicx}
\usepackage{multicol}
\usepackage{multirow}
\usepackage{booktabs}
\usepackage{xcolor}
\usepackage{datapie}
\usepackage{subfig}
\usepackage{xcolor}
\usepackage{tikz}
\usepackage[utf8]{inputenc}
\IEEEoverridecommandlockouts                              

\title{\LARGE \bf{RoadText-1K: Text Detection \& Recognition Dataset for Driving Videos}}

\author{Sangeeth Reddy$^{1}$, Minesh Mathew$^{1}$, Lluis Gomez$^{2}$, Marçal Rusinol$^{2}$, Dimosthenis Karatzas$^{2}$ and C.V. Jawahar$^{1}$
\thanks{$^{1}$The authors are with Center for Visual Information Technology (CVIT), IIIT Hyderabad, India.
        {\tt\small {sangeeth.battu@research.iiit.ac.in}}}
\thanks{$^{2}$The authors are with Computer Vision Center (CVC), UAB, Spain.
        {\tt\small {lgomez@cvc.uab.es}}}
}
\begin{document}

\maketitle
\thispagestyle{empty}
\pagestyle{empty}

\begin{abstract}
Perceiving text is crucial to understand semantics of outdoor scenes and hence is a critical requirement to build intelligent systems for driver assistance and self-driving. Most of the existing datasets for text detection and recognition comprise still images and are mostly compiled keeping text in mind. This paper introduces a new "RoadText-1K" dataset for text in driving videos. The dataset is 20 times larger than the existing largest dataset for text in videos. Our dataset comprises 1000 video clips of driving without any bias towards text and with annotations for text bounding boxes and transcriptions in every frame. State of the art methods for text detection, recognition and tracking are evaluated on the new dataset and the results signify the challenges in unconstrained driving videos compared to existing datasets. This suggests that \textit{RoadText-1K} is suited for research and development of reading systems, robust enough to be incorporated into more complex downstream tasks like driver assistance and self-driving. The dataset can be found at \url{http://cvit.iiit.ac.in/research/projects/cvit-projects/roadtext-1k}


\end{abstract}
\section{Introduction}

\begin{figure}[t]
 \includegraphics[width=\linewidth]{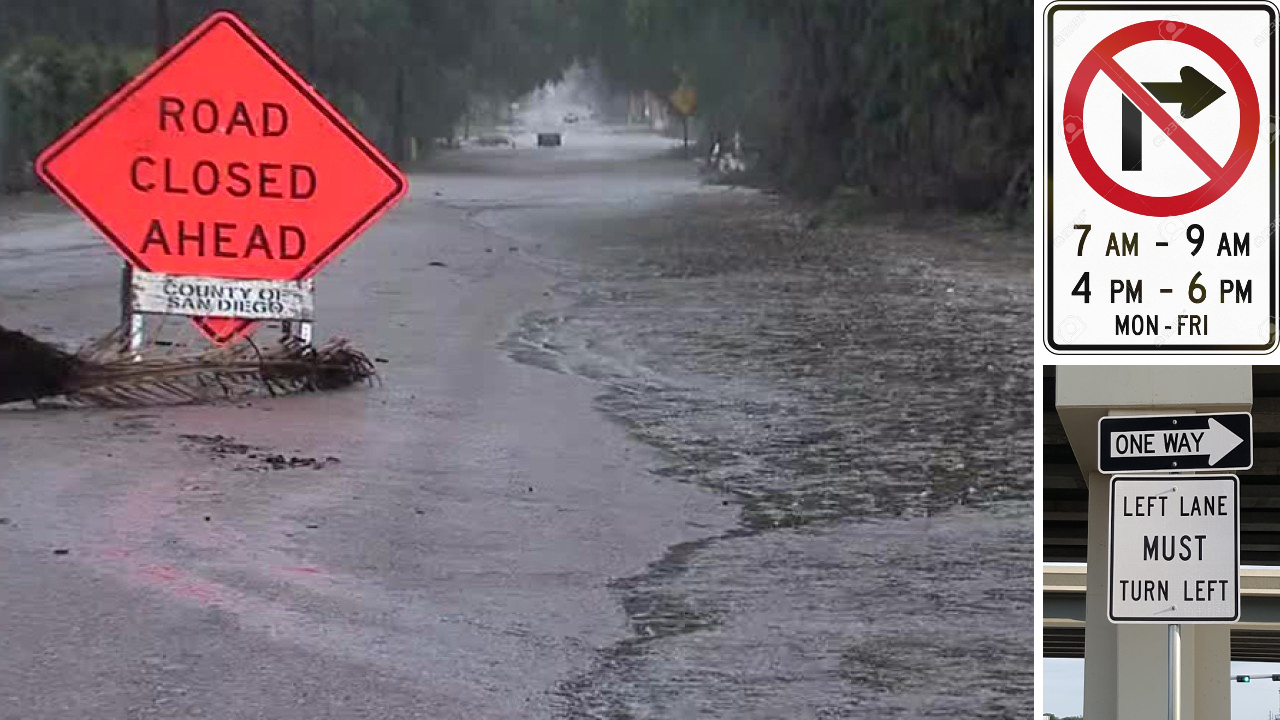}
\vspace{-1.5em}
\setlength{\belowcaptionskip}{-1em}
\caption{\small {Examples showing the importance of text for self-driving and advanced driver assistance systems. Navigation systems in the case of scene on the left, without cognizance of text would be taking the road while it is not supposed to. Similarly the scene at the top-right shows no right turn, but the restriction is only for the timings mentioned. The image on the bottom-right has one way sign to the right but has conflicting message through text.}}\label{fig1}
\end{figure}

Recently, advanced driver assistance and self driving systems have become an active research area. Current driver assistance systems and self-driving approaches mostly disregard textual information on road although text is a medium for conveying crucial information to human drivers. Autonomous navigation systems rely on information from maps, sensory and visual feed~\cite{DBLP:journals/corr/BojarskiTDFFGJM16,DBLP:conf/ivs/Chen017,DBLP:journals/tiv/PadenCYYF16} for route planning and safe navigation on road. Most of the existing driving datasets \cite{kittiDataset,IDDdataset,apollo} have pixel level annotations for objects and other semantic aspects, but do not have text annotated. However, in a real driving environment, it is very common to have unanticipated situations on road leading to interim diversions and detours from regular regime. Generally in such cases text warning boards are used as medium of communication to the driver. Such situations make it necessary for the system to understand the text and act accordingly. 
A few of such instances where cognizance of text is of paramount importance is depicted in the Fig.~\ref{fig1}. 
A new dataset for this purpose is introduced motivating to have systems that are capable of detecting and recognizing text precisely on road, equipping the relevant systems with textual information in real time.

Scene Text detection and recognition has drawn a lot of interest in the computer vision community for its applications in varied domains, ranging from aiding visually impaired individuals to image search and retrieval. With the advent of deep learning and abundance in digital data there has been considerable progress in scene text detection and recognition. Though the application of scene text in images is well explored for retrieval~\cite{DBLP:journals/tmm/KaraogluTGS17, DBLP:conf/iccv/MishraAJ13, DBLP:conf/eccv/GomezMRK18}, fine grained classification of products~\cite{Karaoglu17,DBLP:journals/access/BaiYLXL18} and businesses~\cite{DBLP:journals/tmm/KaraogluTGS17,DBLP:journals/access/BaiYLXL18}, text translation~\cite{DBLP:conf/icra/ShiX05} and tools for the visually impaired~\cite{Rong2016GuidedTS}, it has not been completely incorporated into driver assistance systems or self-driving, except for sign board detection~\cite{DBLP:journals/tits/WuCY05,DBLP:journals/tits/YuanXW17}.
\begin{figure*}[!htb]
\minipage{0.245\textwidth}
  \includegraphics[width=\linewidth]{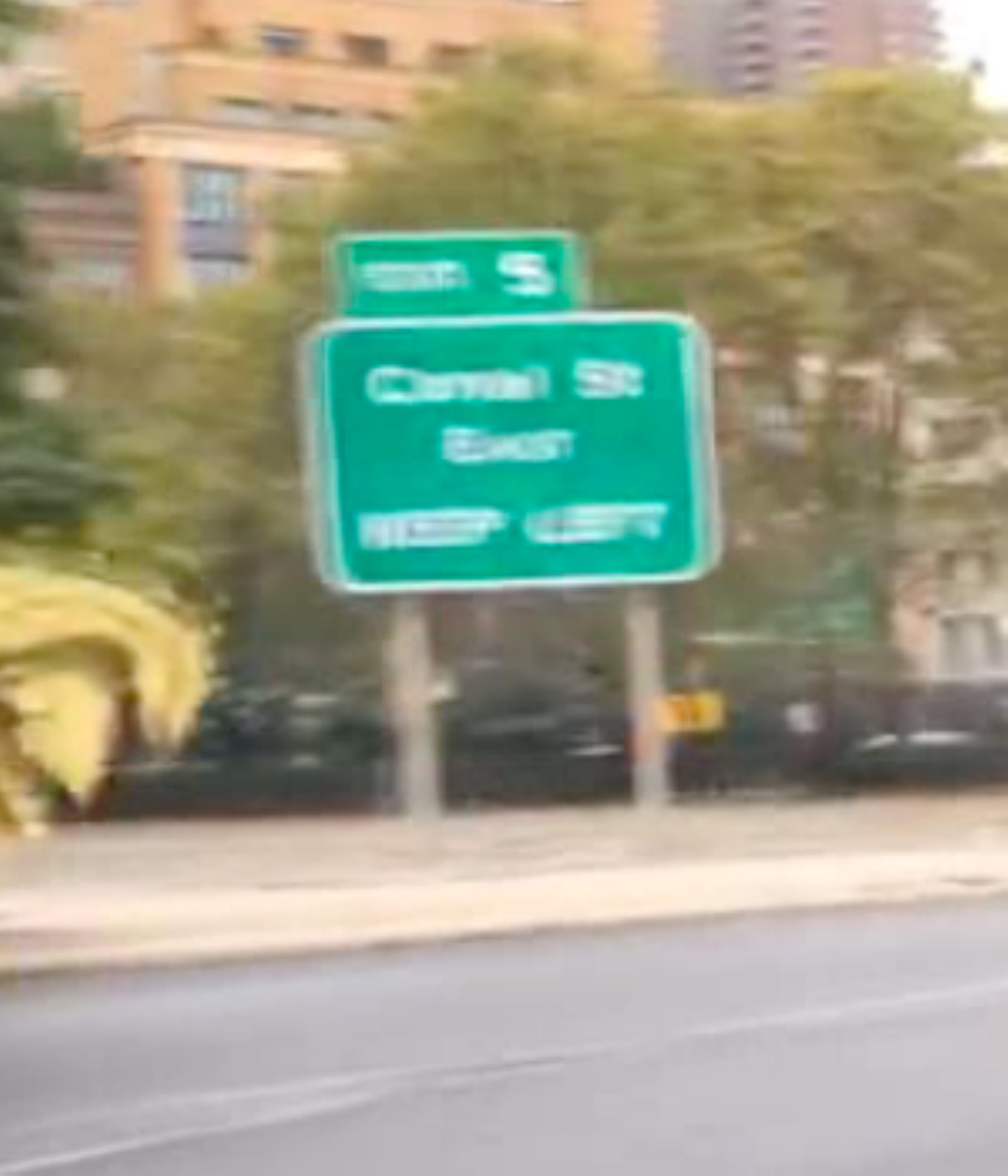}
  \centering\\(a) Motion blur
\endminipage\hfill
\minipage{0.245\textwidth}
  \includegraphics[width=\linewidth]{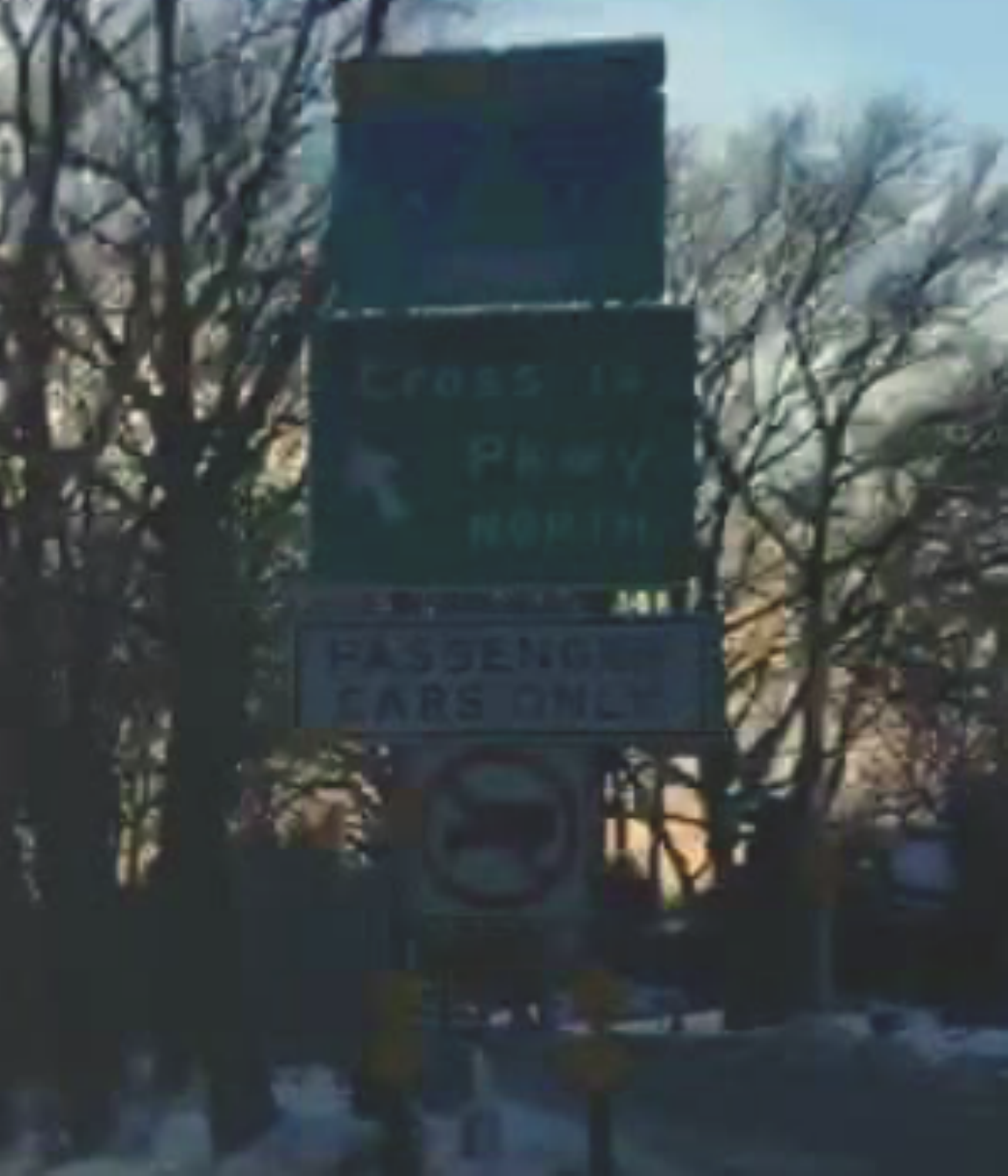}\centering\\(b) Low contrast
\endminipage\hfill
\minipage{0.245\textwidth}%
  \includegraphics[width=\linewidth]{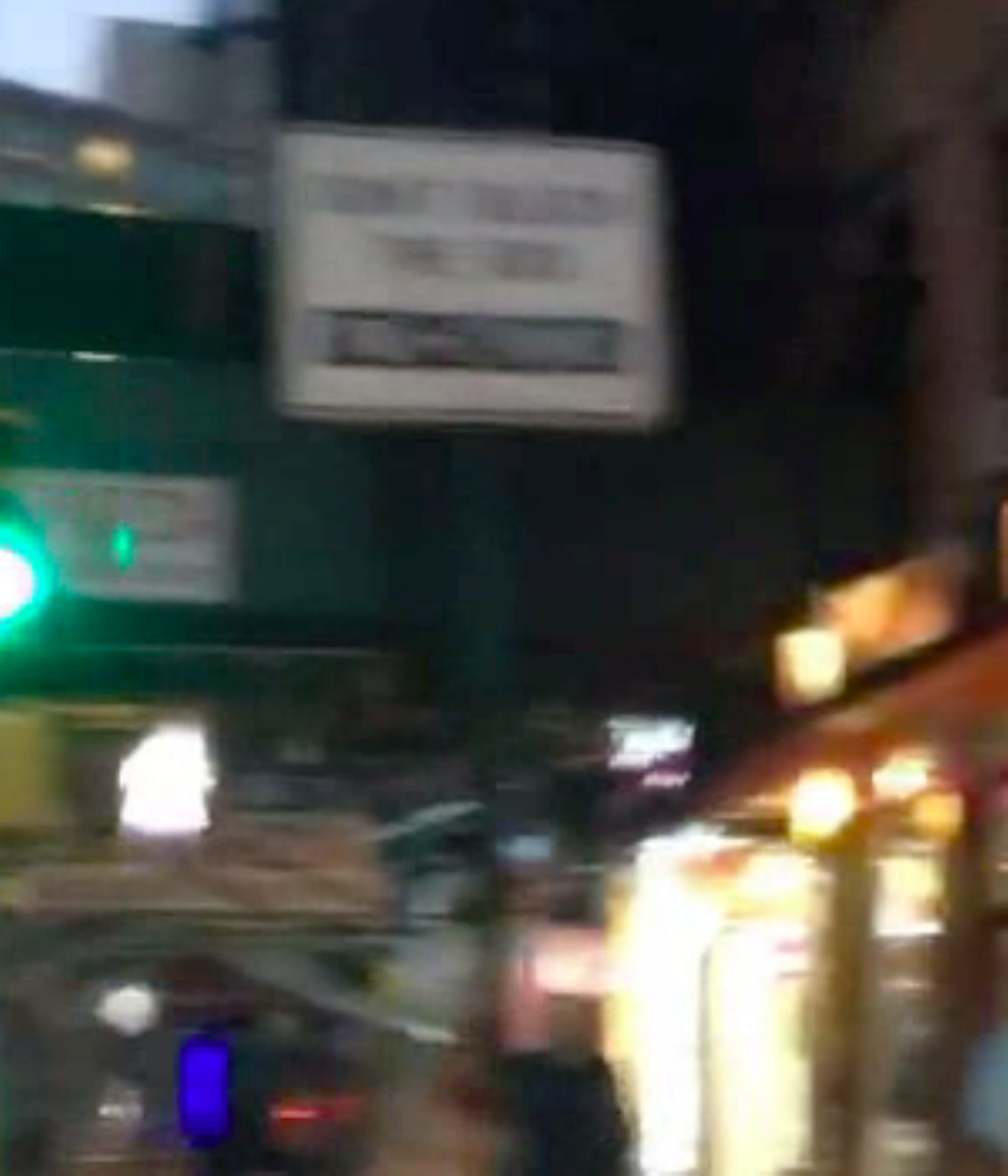}\centering\\(c) Out-of-focus
\endminipage\hfill
\minipage{0.245\textwidth}%
  \includegraphics[width=\linewidth]{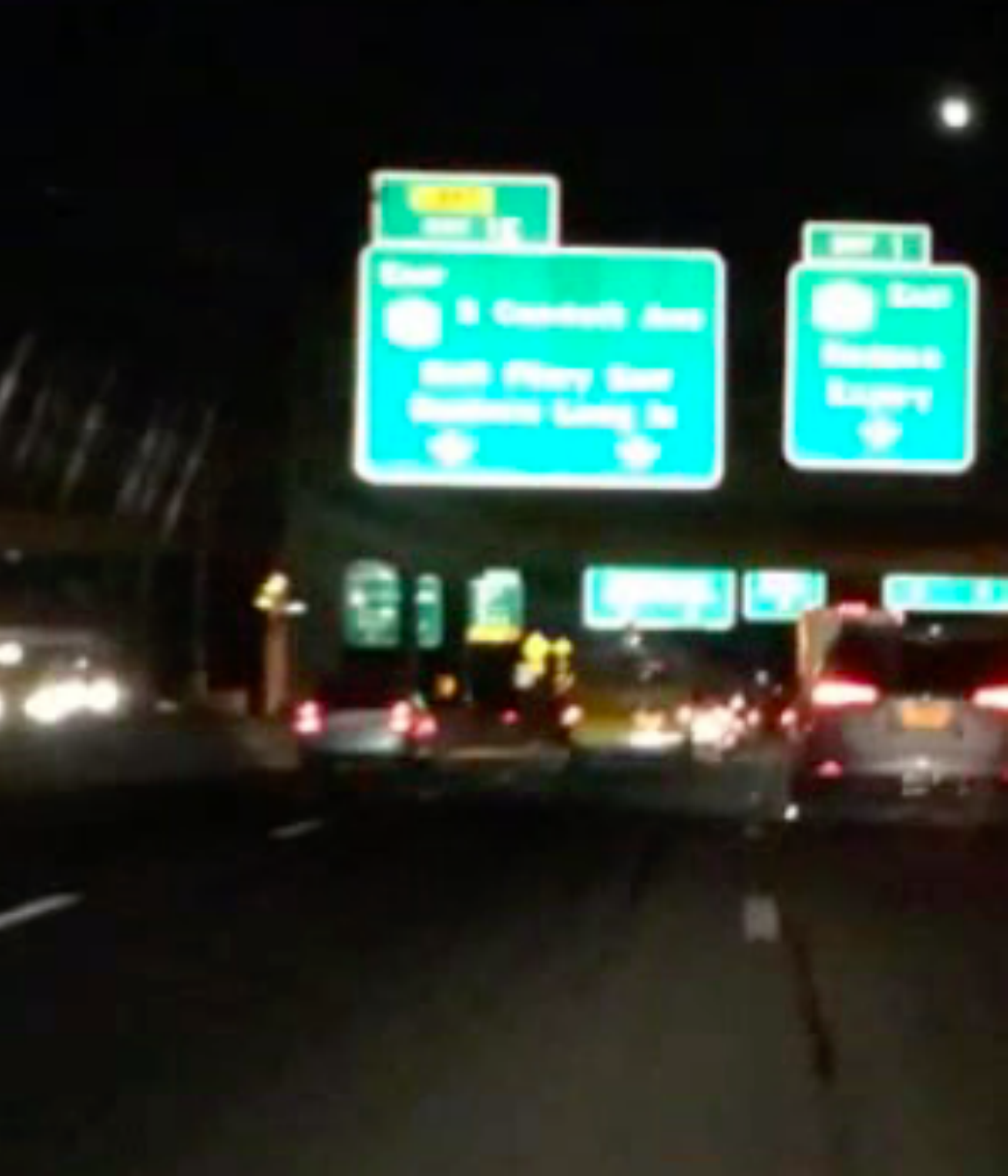}\centering\\(d) Glare
\endminipage
\setlength{\belowcaptionskip}{-1em}
\caption{Frames from RoadText-1K illustrating various challenges/artefacts often encountered in driving videos.}\label{fig2}
\end{figure*}

Most of the recent advances in text detection and recognition in natural scenes~\cite{introduction:1_1,introduction:1_2,introduction:1_3,introduction:1_4,intro:1_5} deal with only static frames. It is often very challenging to detect and recognize text in video frames due to various factors like blur, out-of-focus and other artifacts/distortions as depicted in Fig.~\ref{fig2}. 
There has been recent interest \cite{intro:2_2,intro:2_4} in community for extending text detection and recognition to videos and there are a handful of datasets~\cite{DBLP:journals/pami/TianYSH18,DBLP:conf/icip/MinettoTCLS11, DBLP:conf/wacv/NguyenWB14} that support research in this domain. Existing video text datasets and \textsc{icdar} 2013, Robust Reading Challenge~\cite{DBLP:conf/icdar/KaratzasSUIBMMMAH13} are text centric and curated specifically for this purpose. In the case for autonomous navigation, text present in driving environment is widely spread on scene and camera need not necessarily be centered on text. Adding to it, camera movement incorporates artifacts like motion blur. 
Hence the techniques built on existing datasets which have focused and centered text, are quite not a match for real world applications such as driver assistance and self-driving systems. 
Our work intends to contribute a dataset RoadText-1K to the community for developing and testing systems that can fare in realistic settings.

Contributions of this work are, 
\begin{itemize}

\item We create a large scale, diverse and unconstrained dataset of driving videos with dense annotations of text location and transcription. The proposed dataset is 20 times the existing largest dataset~\cite{DBLP:conf/icdar/KaratzasSUIBMMMAH13}. 
\item License plates in the videos are separately tagged to distinguish them from other text instances. This would make the data useful for the problem of license plate detection and recognition.
\item We evaluate current state of the art techniques for scene text detection, recognition, tracking and provide a thorough analysis of performance on this dataset.
\end{itemize}

Rest of the paper is structured as follows, Section II discusses the related work concerning text detection and recognition in images and videos. Section III details how new dataset is compiled, annotated and presents statistics, comparative analysis. In Section IV we present results of state-of-the-art text detection, recognition and tracking methods on the new dataset. Finally, Section V presents the conclusion and thoughts for future work.
\section{Related Work}
In this section we  discuss various related works concerning detection and recognition of text in images and videos, followed by a discussion of works which make use of text in scenes for other applications and downstream tasks.
\subsection{ Text detection and recognition in scene images}
Text detection and recognition has been an interesting field of research for a long time. 
With the advent of deep learning and success of Convolutional Neural Networks (\textsc{cnn}), most of the recent text detection and recognition methods have been exploiting their share of utility from these learning models.
Strong performance of \textsc{cnn} for object detection tasks has motivated the community to use them for text detection. Text being treated as an object in the image, various methods \cite{EE:1}\cite{DBLP:conf/cvpr/ZhouYWWZHL17}\cite{DBLP:conf/cvpr/LiuLYCQY18} based on object detection architectures as backbone have been proposed and have improved the performance considerably. On the other hand, most recent methods for text recognition i.e, the task of transcribing text in a localized region in an image, treat the problem as a sequence to sequence translation problem and primarily rely on a Convolutional Recurrent Neural Network (\textsc{crnn}) style architecture~\cite{DBLP:journals/pami/ShiBY17} or an encoder-decoder approach~\cite{bshi2018aster}.


\subsection{Text detection and recognition in videos}
In general, video frames particularly of an outdoor video involving motion, is subject to various artefacts like motion blur and defocus 
Methods designed for still images, may fail to obtain reliable detection and recognition results when applied to frames of a video. On the other hand a text instance appears in multiple frames in a video and this temporal redundancy can be of help in improving the recognition.
Exploiting the redundancy and correlation of textual features across temporal domain is expected to improve the detection and recognition results compared to single frame level methods. Various video text detection methods explore this strategy \cite{DBLP:conf/icpr/GomezK14,DBLP:conf/icip/ShivakumaraLWL14,DBLP:conf/icdar/TanakaG07,DBLP:journals/tmm/WuSLT15,DBLP:journals/tip/YangYPTZZY17} by techniques such as tracking, multi-frame integration and spatio-temporal analysis. Yin et al.~\cite{DBLP:journals/tip/YinZTL16} summarizes text detection, tracking and recognition methods in video and their challenges.
\begin{figure*}[h]
\begin{center}
\includegraphics[width=.325\textwidth]{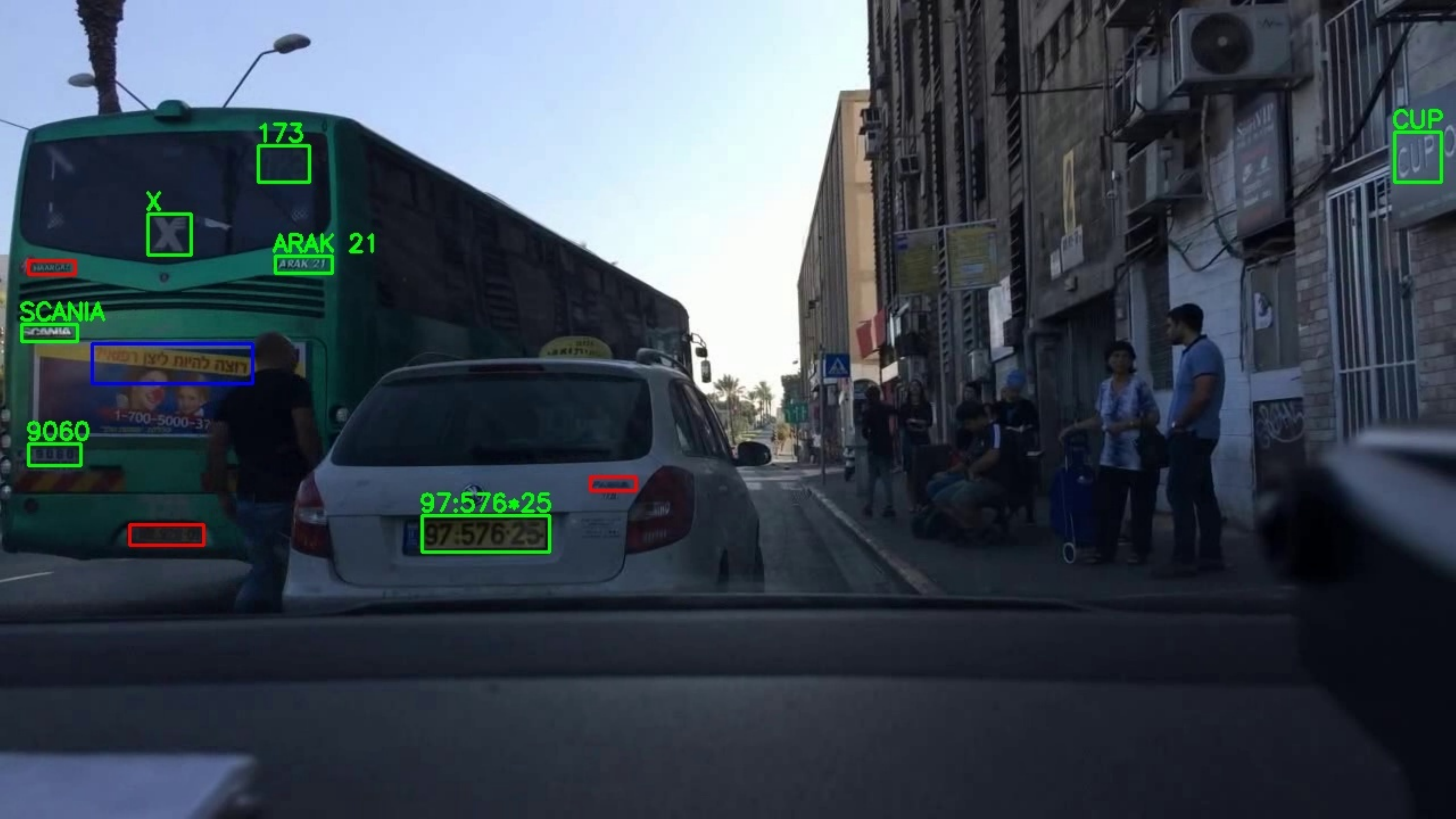}
\includegraphics[width=.325\textwidth]{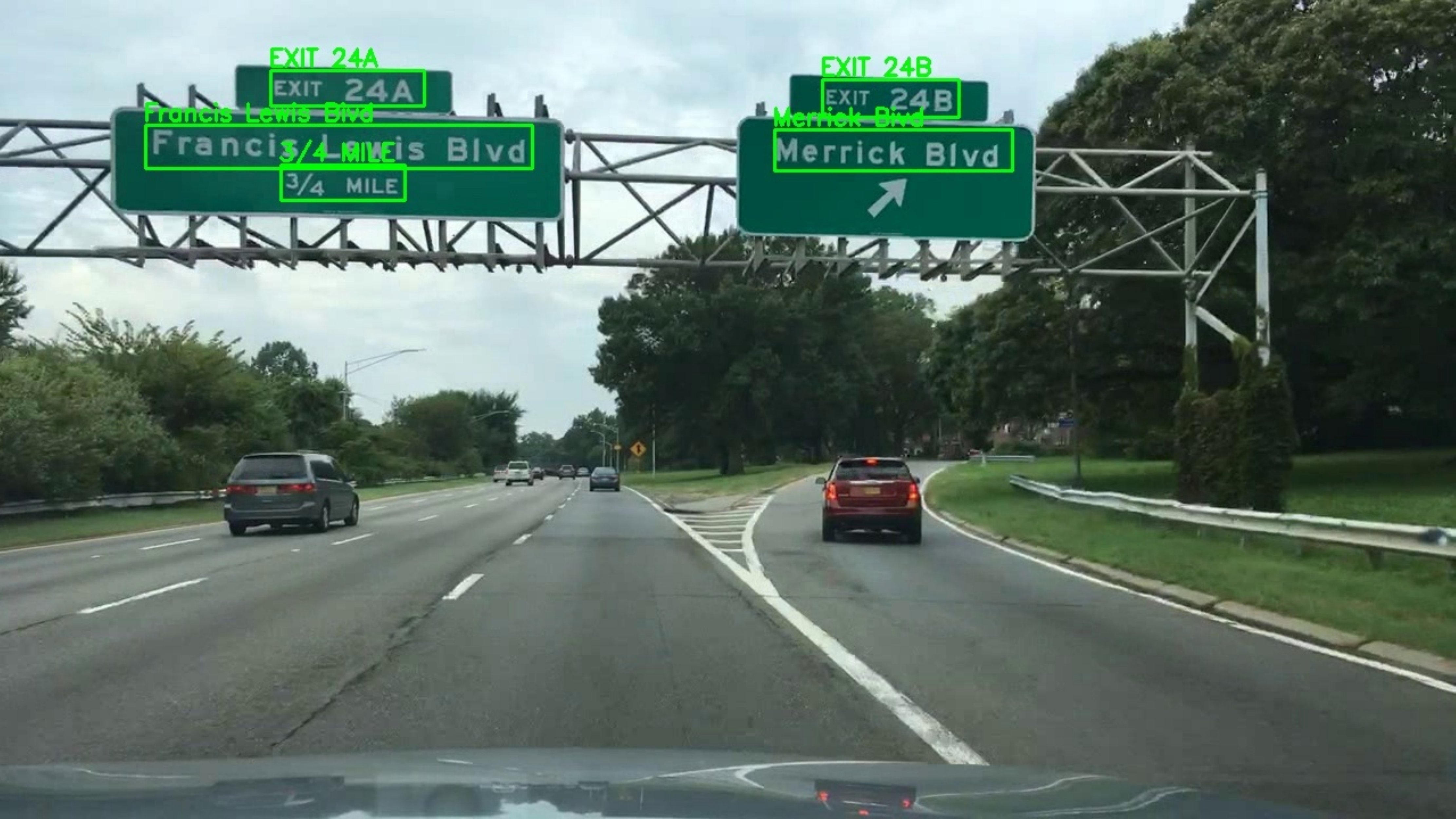}
\includegraphics[width=.325\textwidth]{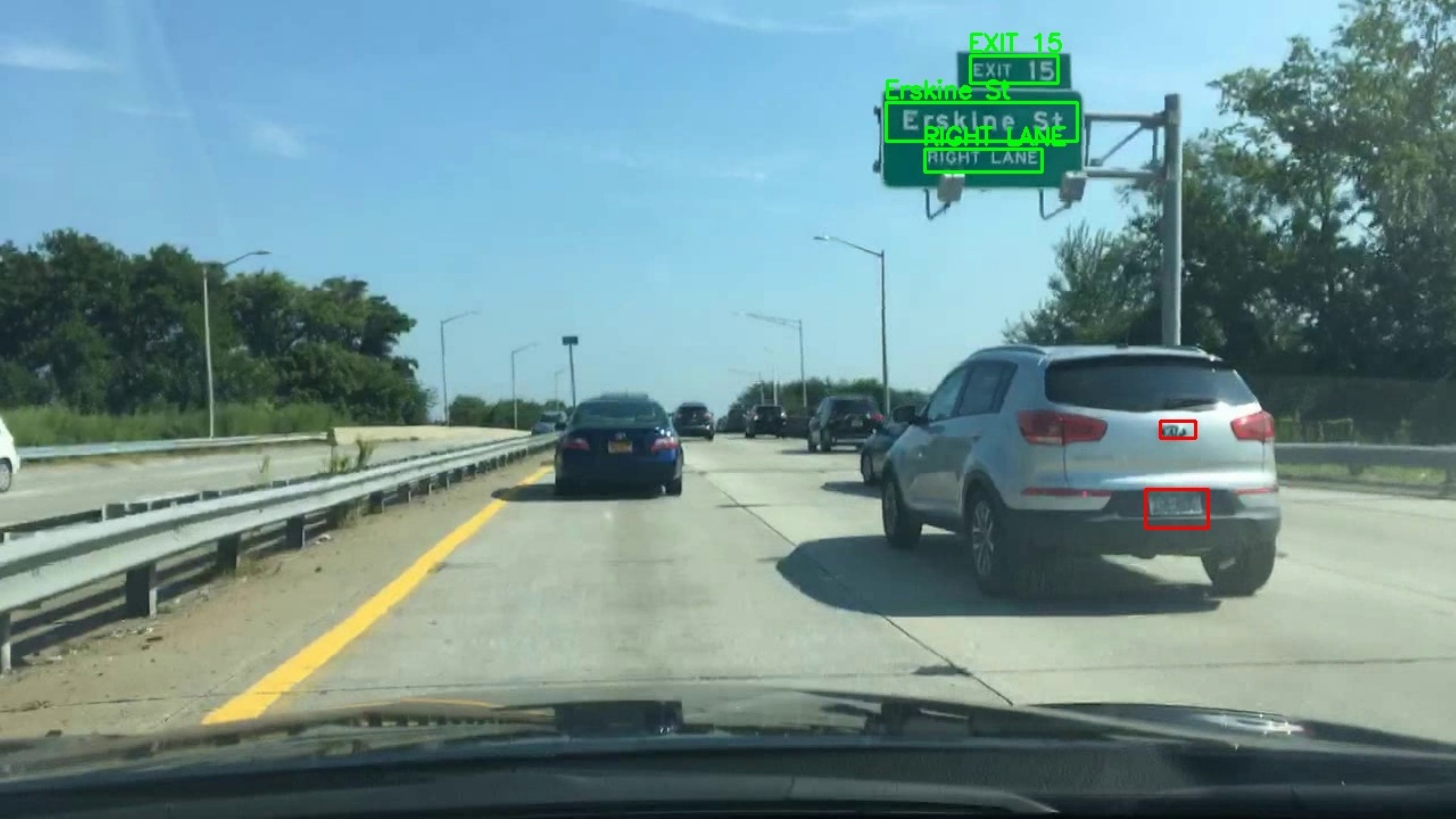}
\setlength{\belowcaptionskip}{-2em}
\caption{Example frames from clips in RoadText-1K with text location and transcription annotations overlaid. Boxes in green correspond to English text, blue represent non English and red represent illegible text.
}
\label{fig3}
\end{center}
\end{figure*}

Once the text regions are tracked using various tracking methods, one of these two techniques are typically employed for better recognition:(1)~\textit{(selection technique)} by selecting best text instances from tracked text regions as proposed in \cite{DBLP:conf/icpr/ShiratoriGK06} \cite{DBLP:conf/icdar/TanakaG07}, and (2)~\textit{(fusion technique)} by combining consecutive recognition results. Rong et al. \cite{DBLP:conf/icmcs/RongYYT14} fuse multiple recognition results of same text region and final result is considered by either majority voting or fusing based on confidence or other metrics. Nevertheless these approaches are computationally expensive since the detection and recognition models essentially run on every frame.

Wang et al. in~\cite{DBLP:conf/icdar/WangJYZLFWL17} propose a method for end-to-end text recognition utilizing correspondences across multiple frames. They use spatio-temporal redundancy for text detection and edit distance for recognized text to fuse results from multiple frames. Recent work from Cheng et al.~\cite{2:1}, propose a scene text spotting framework in video with a spatiao-temporal detector and discriminative tracker to recognize text once per track by picking the best instance using a quality scoring mechanism. In section V we evaluate a similar strategy for tracking on the proposed RoadText-1K dataset. 



\begin{figure*}[t]
\setlength{\belowcaptionskip}{-1em}
\minipage{0.32\textwidth}
  \includegraphics[width=\linewidth]{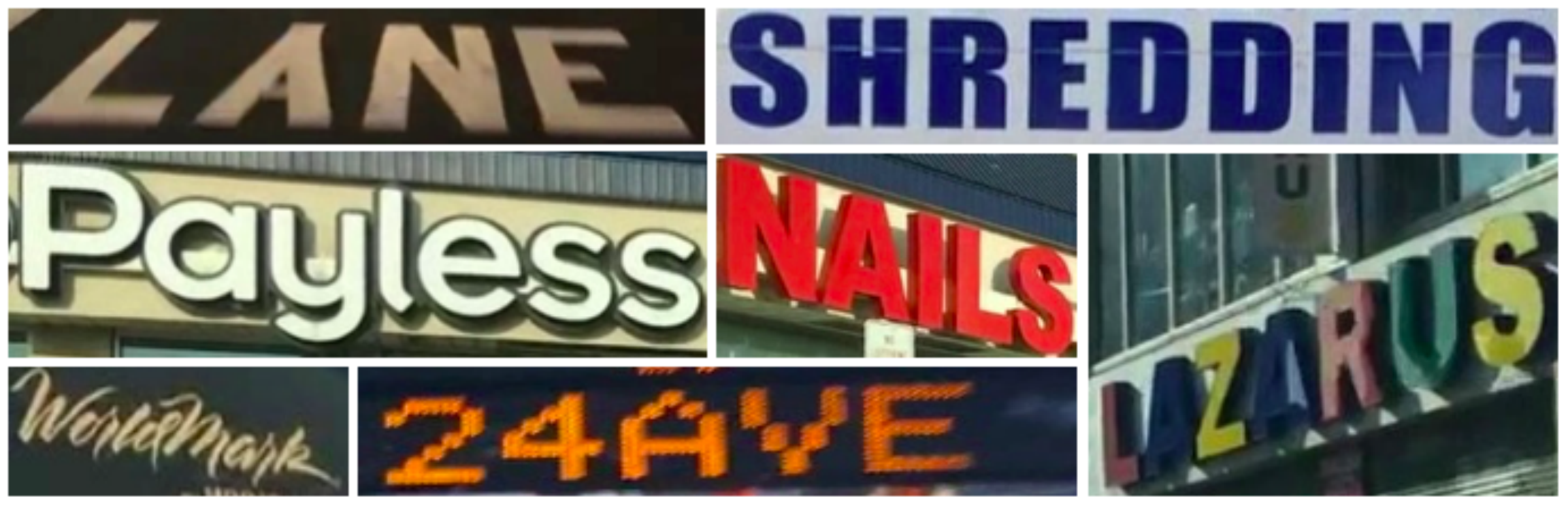}\centering\\(a) English
\endminipage\hfill
\minipage{0.32\textwidth}
  \includegraphics[width=\linewidth]{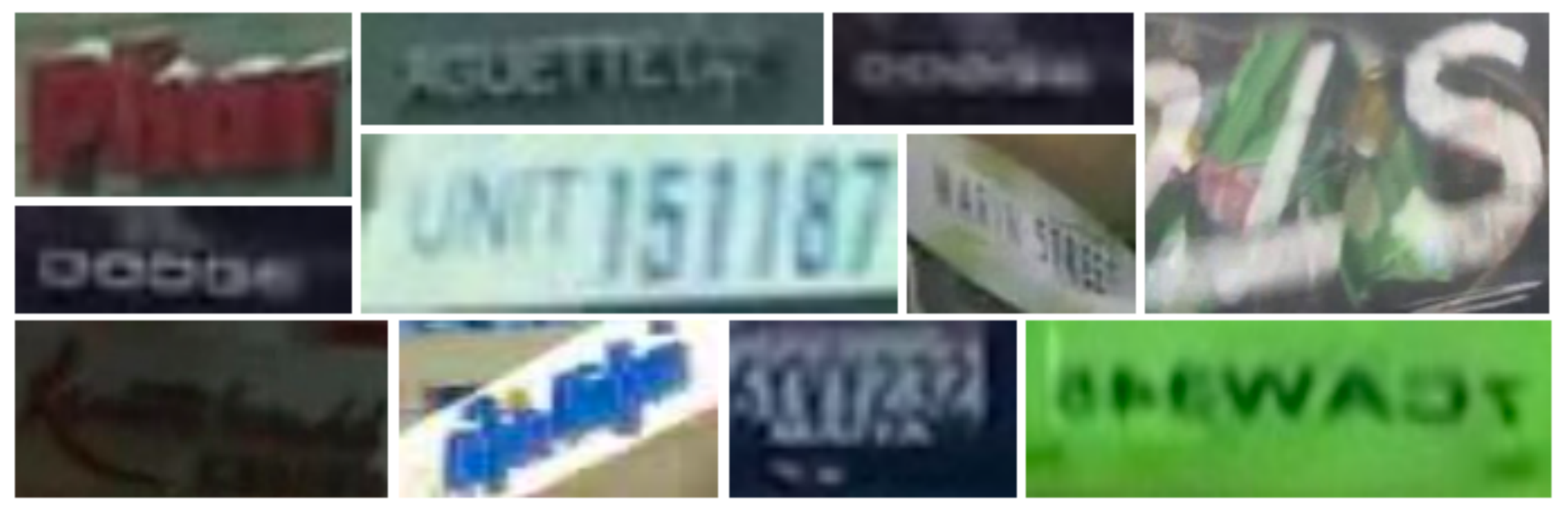}\centering\\(b) Illegible
\endminipage\hfill
\minipage{0.32\textwidth}%
  \includegraphics[width=\linewidth]{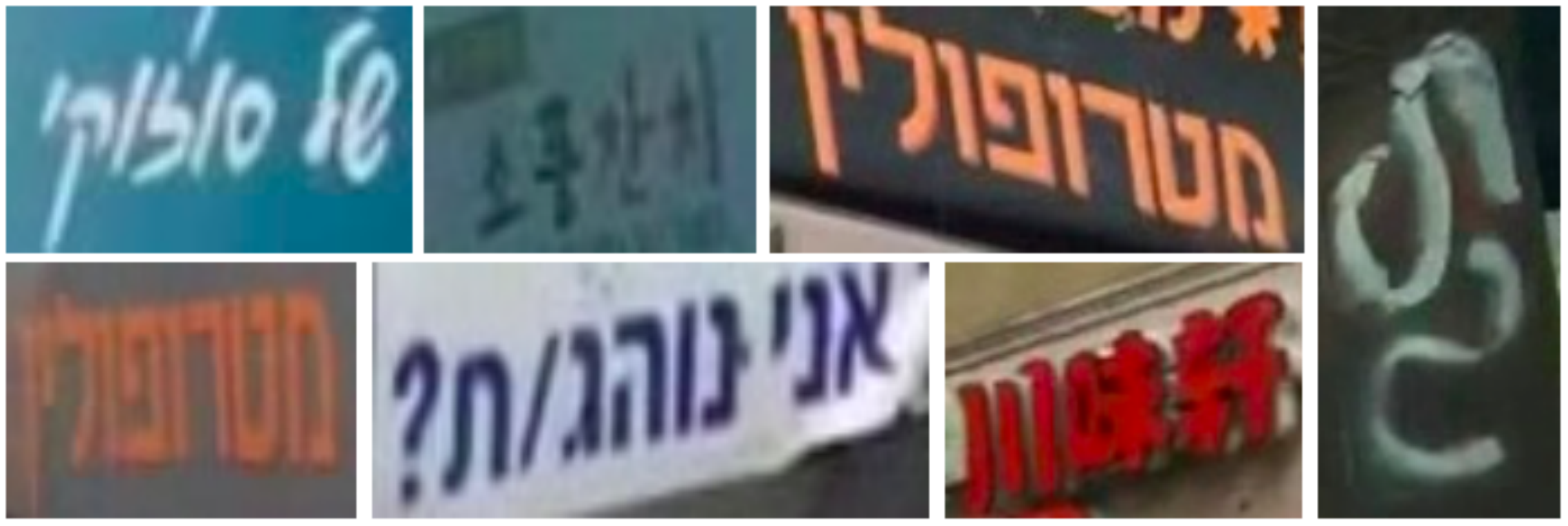}\centering\\ (c) Non English
\endminipage
\caption{\small Samples of text instances for each text category}\label{fig:5}
\vspace{-1em}
\end{figure*}
\subsection{Scene text understanding for navigation}
Autonomous navigation systems highly rely upon visual and sensory inputs. A camera feed is used commonly for semantic segmentation or classification of scene into drivable area, sidewalk, vegetation, sky etc. However the text present in the scene which conveys high level semantic information, is often ignored.

Case et al.~\cite{DBLP:conf/icra/CaseSCN11} propose to use text present in the scene to generate automatic semantic labels during robotic mapping of an indoor environment. Work by Wu et al.~\cite{DBLP:journals/tits/WuCY05} deals with text detection on road signs with application to driver assistance systems. In this work the sign boards are first detected in each frame of the video and then features like color, edges and texture are used for detection of text on the sign boards. 
Gonzalez, et al.~\cite{DBLP:journals/tits/GonzalezBT14} attempts to extract the text present in road panels of street images as an application to Intelligent Transportation Systems (\textsc{its}). 
A character and word recognizer were used for recognition and to improve the recognition, a Web Map Service is used to restrict the dictionary size to a limited geographical area.
Shi et al.~\cite{DBLP:journals/tits/GreenhalghM15} propose to detect and recognize text in traffic signs by using Maximally Stable Extremal Regions (\textsc{mser}), hue, saturation, value color thresholding with constraints upon temporal and structural information for text region detection. Individual text characters are detected as \textsc{mser} and grouped into lines followed by an Optical Character Recognition (\textsc{ocr}) module. Zhou et al.~\cite{DBLP:conf/cvpr/ZhouYWWZHL17} detects text-based traffic signs using a convolutional network as region proposal network and another neural network for final classification of text regions from  the proposals.

Our work goes beyond traffic-sign text detection, and introduces a new large-scale, densely annotated dataset and benchmark for scene text detection and recognition in an unconstrained, realistic driving setup.

\section{RoadText-1K}

We start by explaining how videos for the dataset were selected from an existing driving videos dataset followed by annotation process. Finally, we provide statistics and analysis of the dataset in comparison with existing datasets for text in videos

\subsection{Videos}
The 10 second long video clips in our dataset are sampled from BDD100K~\cite{bdd:1}, which contain 100K driving videos. Each video in BDD100K is about 40 seconds long, 720p and 30 fps collected with diversity across locations in the United States, weather conditions including sunny, overcast, and rainy, as well as different times of day including day and night. Videos in BDD100K are collected with an intent to make robust self driving systems and were not specifically collected for the task of text detection or any aligned task. This makes BDD100K video database an ideal source of raw videos that are diverse and unconstrained without any specific bias towards the problem of understanding text in images/videos.

We ran an off-the-shelf text detector~\cite{DBLP:conf/cvpr/ZhouYWWZHL17} on frames of the videos in BDD100K to shortlist videos which have considerable number of text instances. In the next step, 1000, 10 seconds long video clips are handpicked from this shortlisted collection. We performed this step manually to  make sure that the clips selected are diverse in terms of scenes, and have good number of English text. We split the 1000 videos in the dataset randomly as 700 for training and 300 for test.
 

\subsection{Annotation of text instances}
\begin{figure}[htbp]
\centering
\includegraphics[width=.48\textwidth]{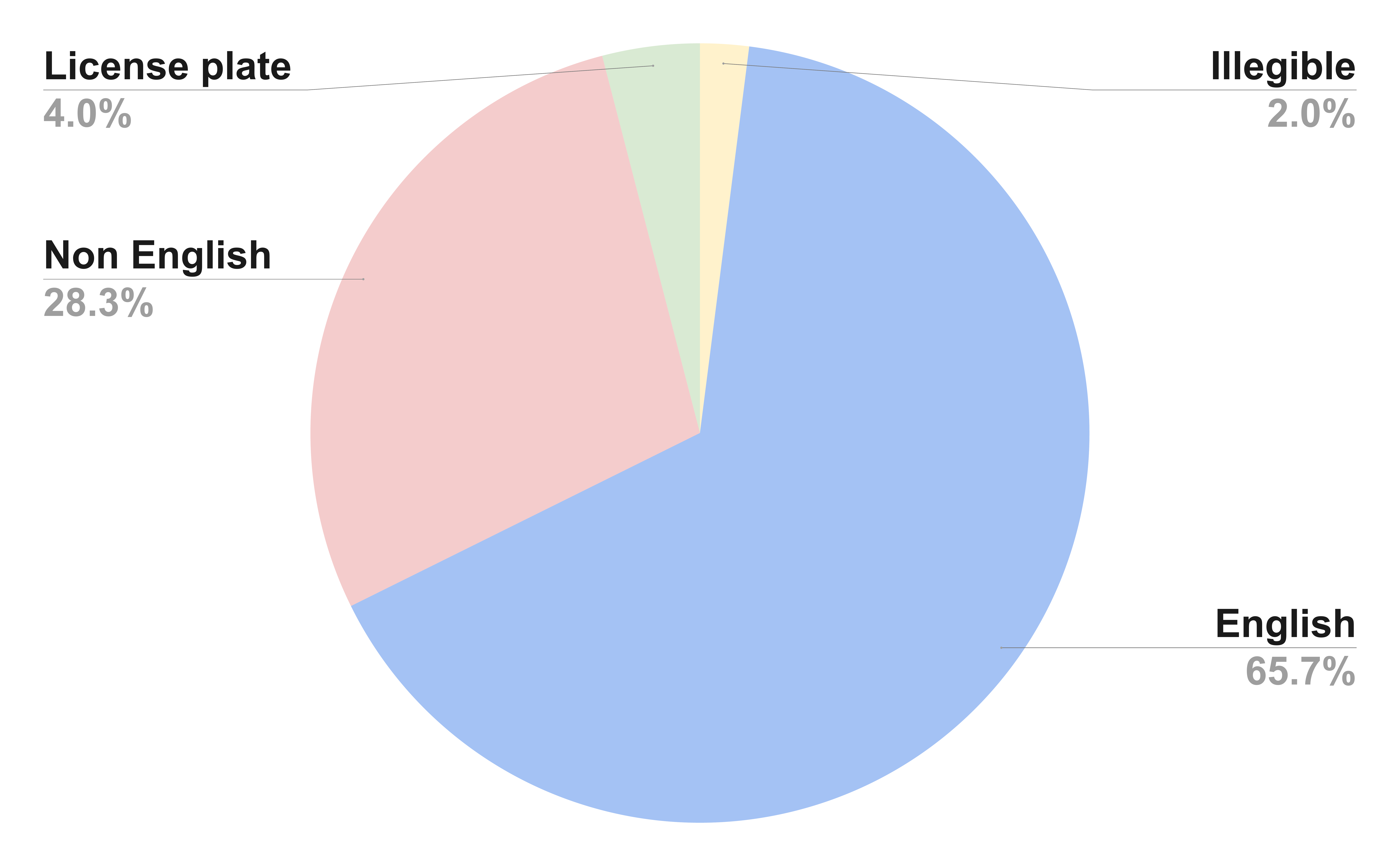}
\setlength{\belowcaptionskip}{-1em}
\vspace{-2.0em}
\caption{\small Category distribution of text instances in RoadText-1K.}
\label{fig7}
\end{figure}

The annotation of text instances in every frame of the video clips involved a two stage process. In the first stage the annotators were asked to bound the text instances by a bounding box and to assign each of them with a text type category from the following three categories: (1) English (2) Non English (3) Illegible. In English category we further classify into (a) English or (b) license plate. Distribution of text instances in RoadText-1K based on the text type is shown in Fig.~\ref{fig7}. 

Unlike in the case of most scene text datasets we annotate text lines rather than annotating each `word' (split at spaces).
This approach makes annotation much faster and also avoids the ambiguity in deciding how to split text into 'words' in certain cases where numbers or abbreviations are involved. 
Also when text is annotated at `word' level, it will result in many smaller tokens like a single full stop, a single `a' or a single digit which are typically difficult to detect and recognize. Moreover the recent text recognition methods using Connectionist Temporal Classification (\textsc{ctc})~\cite{DBLP:conf/icml/GravesFGS06} or an encoder-decoder framework doesn't require the text line to be split into words or sub-words. Also transcribing a text line at once lets the transcription model access more context, and can benefit from an implicit language model which is learnt as part of the process~\cite{sabir_lstm_language_model}.

Dense annotation of text in video is a complex and time consuming task, since it involves annotating bounding box for every text instance and at every frame.
The major advantage of video is to have multiple occurrences of same text temporally and with considerably small change in spatial locations depending on the motion.
Hence we used  Scalabel\footnote{\url{https://www.scalabel.ai}}
for bounding box annotation as it enables tracking across frames.
Once a text instance is marked with a bounding box, a track id and category label, the tool aids in tracking the instance across frames and assigns the same track id and category label to other occurrences of the same text instance. The human annotator then reviews the boxes frame by frame and manually adjusts the boxes wherever tracking did not fit the boxes correctly. There were cases where a text instance undergo transition from  illegible/occluded to a legible case or vice-versa. In such cases, initial track is ended from the point where the transition happens and a new track is started, with a new category label. And in cases where a legible text gets partially occluded or goes out of focus for a few frames in between, annotators were directed to continue the track if the occlusion/out-of-focus is not for more than 3 frames. If it takes more than 3 frames, existing track is ended, and a new track with category label as \textit{illegible} is assigned for the time the text instance is occluded/out-of-focus.

In the second stage of annotation, annotators were given videos annotated with the bounding boxes in the first step and were asked to add text transcriptions for tracks which are assigned with \textit{English} and \textit{license plate} category. For this step we used a tool developed internally for this task. The tool loads a video with the bounding box annotations and displays track ids and the category labels assigned in the first step. Once the annotator adds the text transcription to a certain text instance, it is propagated to all occurrences in the same track. Hence the annotator needs to attend to only those tracks which are of \textit{English} or \textit{license plate} category and not assigned with a transcription yet. Examples of frames with both the bounding box and transcription annotations overlaid are shown in Fig.~\ref{fig3}.

\subsection{Statistics and Analysis}
We summarize the overall statistics of RoadText-1K in Table~\ref{t2} along with a comparison of our dataset with existing datasets for text in videos. 
Top 10 frequent words in our dataset are shown in Fig.~\ref{fig_9}(a), we also plot the distribution of words as a function of word lengths in Fig.~\ref{fig_9}(b) . It is observed from the plot that most words present on road are in general of length 7-8 characters and can go till 41 characters. Fig.~\ref{fig_9}(c) shows number of frames as a function of number of text instances appearing in a frame. Although the dataset is collected on roads, at least 50{\%} of the unique text instances are non traffic/road signs.

Existing datasets we compare in Table~\ref{t2} are : Text in Videos~\cite{IC:15}, USTB-VidTEXT~\cite{DBLP:journals/pami/TianYSH18} and Youtube Video Text (YVT)~\cite{DBLP:conf/wacv/NguyenWB14}. Among these, USTB-VidTEXT and YVT mostly contain born digital text (captions and subtitles) in videos sourced from Youtube. Recognition of born digital text is less challenging since they are free from most of the distortions and imaging artifacts and hence even OCRs designed for documents perform quite well on them.
Text in Videos dataset is the largest among these and have 50 egocentric videos containing mostly scene text with dense annotations for bounding boxes and transcriptions. In Fig.~\ref{fig_10} we show the spatial distribution of text in these datasets compared to RoadText-1K. It can be seen that both RoadText-1K and Text in Videos datasets have text instances spread widely across the frame compared to the other two. And the spread in USTB-VidTEXT is very minimal since it only contains subtitles.

\begin{figure}[h]
\begin{center}
\frame{\includegraphics[width=.48\textwidth]{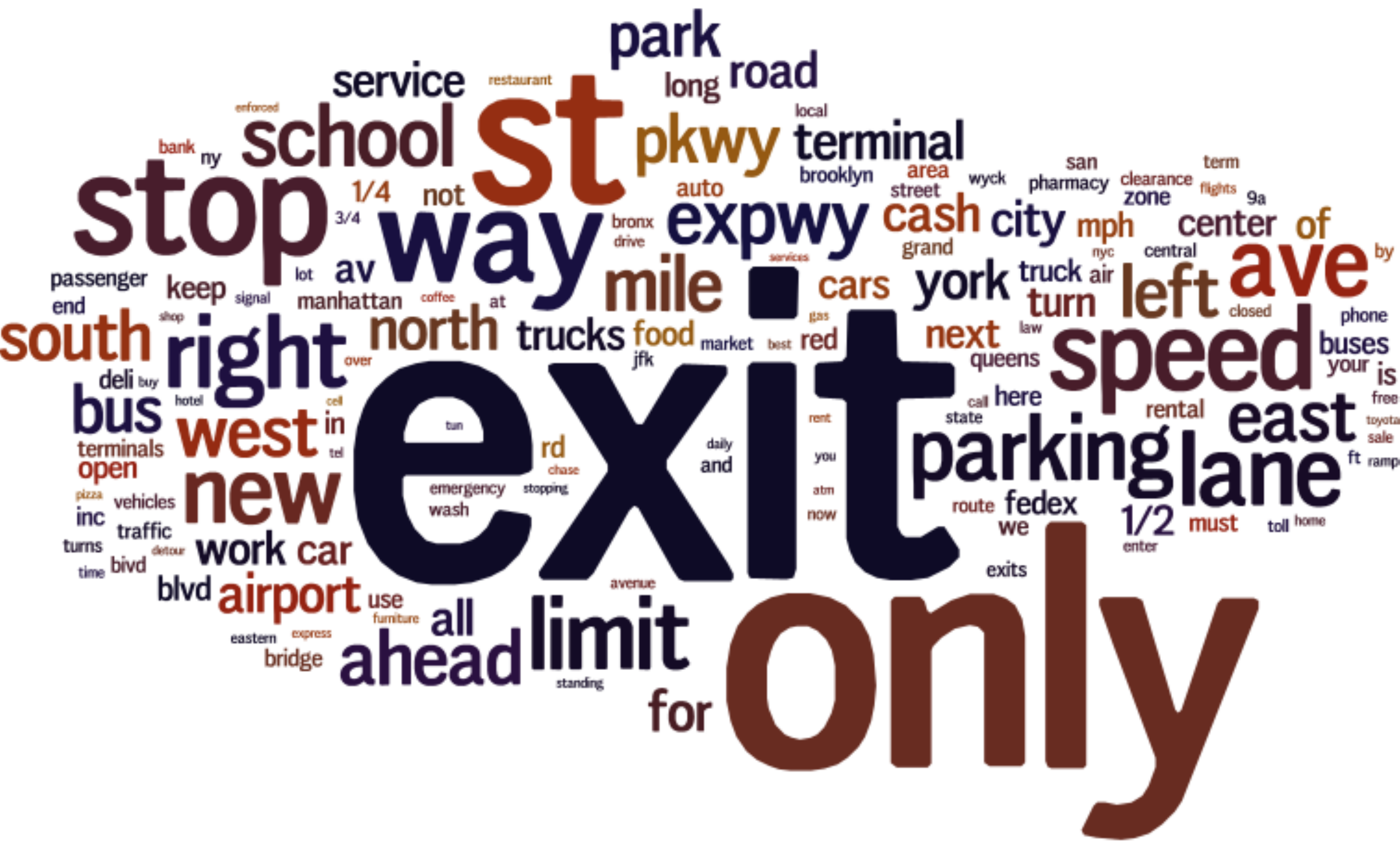}}
\setlength{\belowcaptionskip}{-2em}
\vspace{-1.5em}
\caption{WordCloud of words in RoadText-1K} \label{fig8}
\end{center}
\end{figure}

\begin{table*}[h]
\caption{Comparison of RoadText-1K with existing datasets for text in videos}
\label{t2}
\begin{center}
\scalebox{1.1}{
\begin{tabular}{|c|c|c|c|c|}
\hline
Dataset & Text in Videos~\cite{IC:15} & USTB-VidTEXT~\cite{DBLP:journals/pami/TianYSH18} & YouTube Video Text \cite{DBLP:conf/wacv/NguyenWB14} & RoadText-1K (ours) \\
\hline
Source & Egocentric & Youtube & Youtube & car-mounted \\
\hline
Size (Videos) & 51 & 5 & 30 & \textbf{1000} \\
\hline
Length (Seconds) & varying & varying & 15  & 10 \\
\hline
Resolution & $720\times480$ & $480\times320$ & $1280\times720$ & $1280\times720$ \\
\hline
Annotated Frames & 27,824 & 27,670 & 13,500 & \textbf{300,000} \\
\hline
Total Text Instances & 143,588 & 41,932 & 16,620
& \textbf{1,280,613} \\
\hline 
Text type & Scene Text & Digital (captions) & Scene Text and Digital & Scene Text \\
\hline
Unique Words & 3,563 & 306 & 224 & 8,263 \\
\hline
Avg. text frequency per frame & 5.1 & 1.5 & 1.23 & 4.2 \\
\hline
Avg. Text Track length & 46 & 161 & 72 & 48 \\
\hline
\end{tabular}}
\end{center}
\vspace{-2em}
\end{table*}

\begin{figure*}[!htb]
\minipage{0.32\textwidth}
  \includegraphics[width=\linewidth]{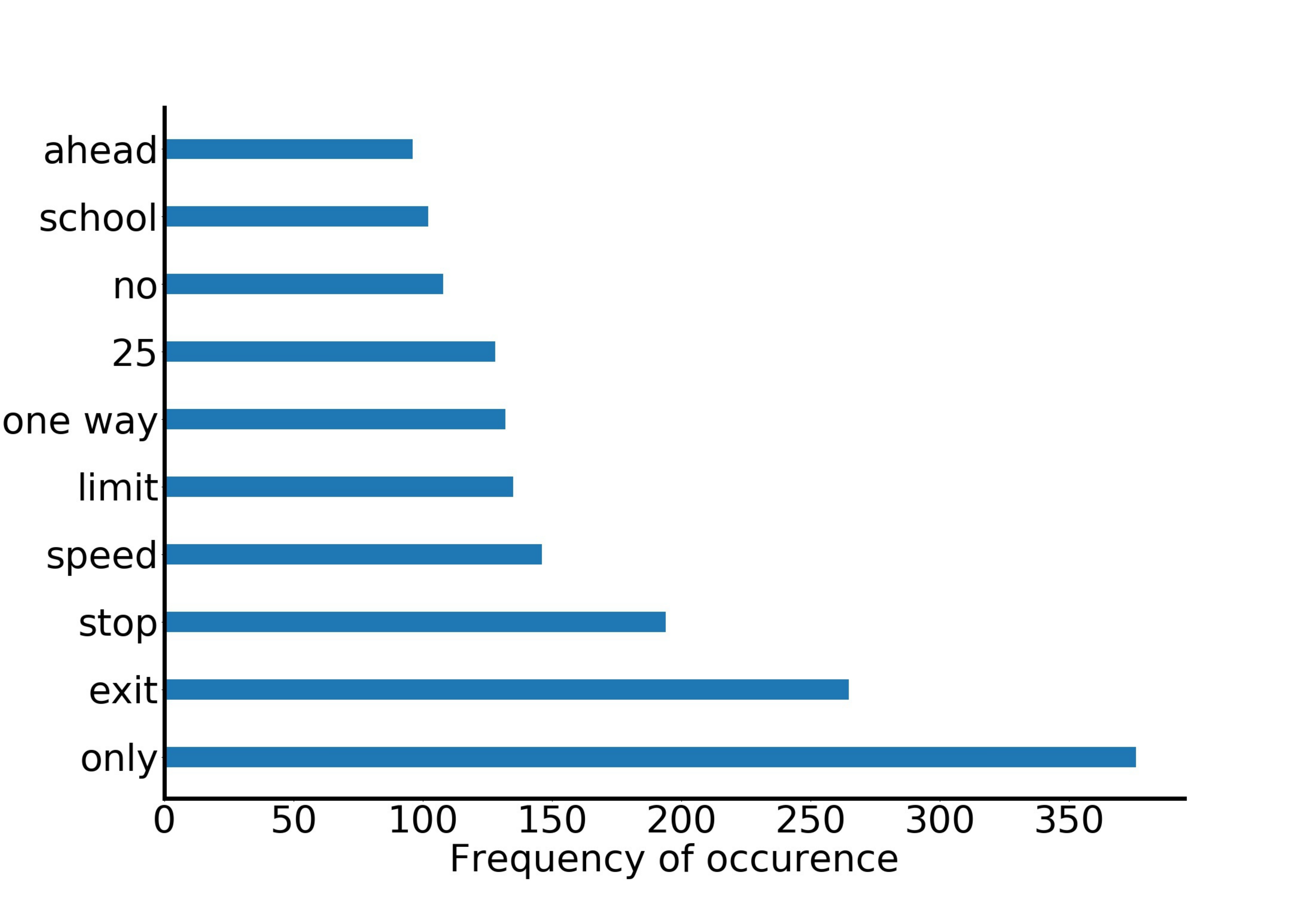}\centering \\\small{(a) Top 10 most occurring words.}
\endminipage\hfill
\minipage{0.32\textwidth}
  \includegraphics[width=\linewidth]{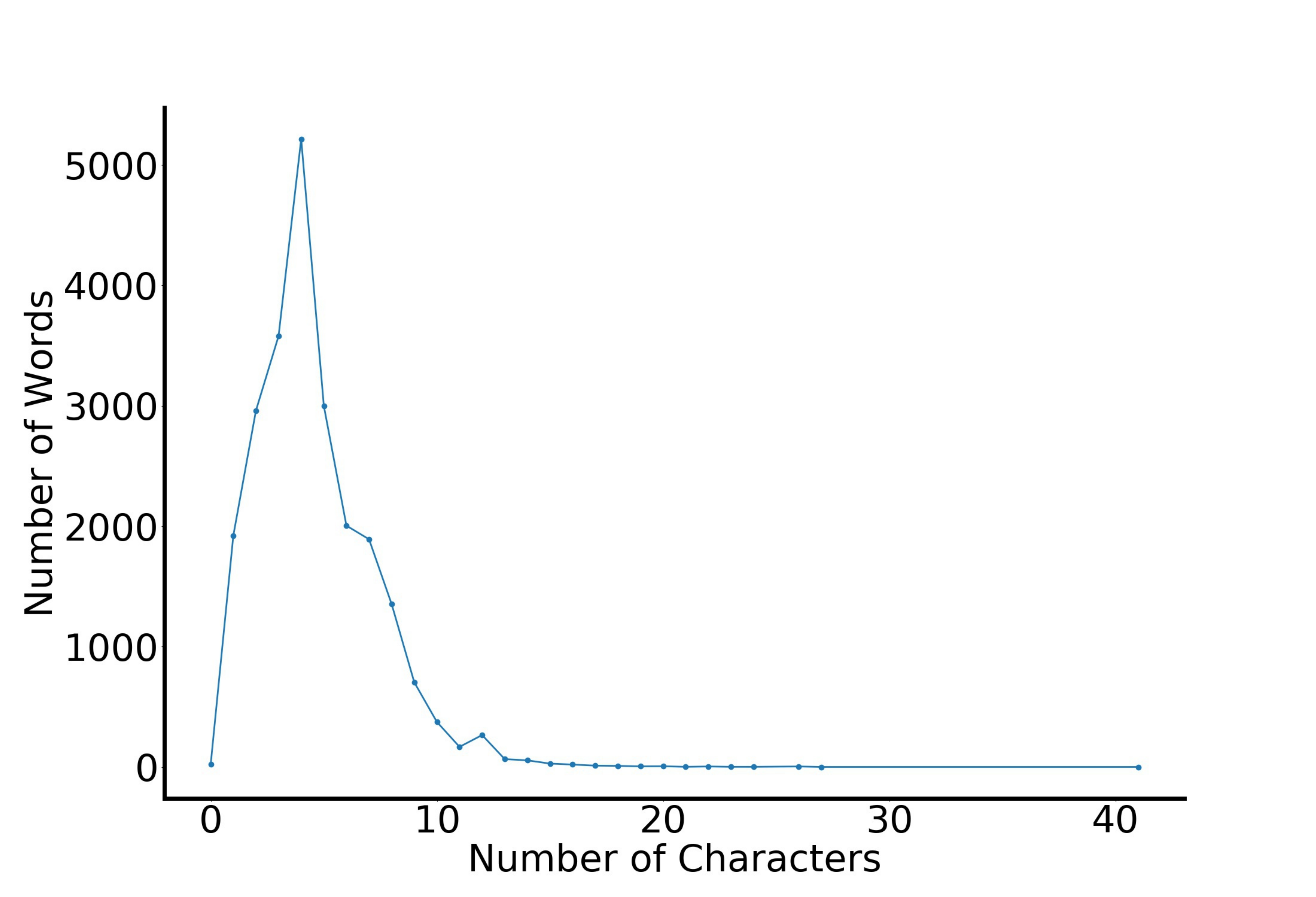}\centering \\ \small{(b) Number of words with a particular length.}
\endminipage\hfill
\minipage{0.32\textwidth}
  \includegraphics[width=\linewidth]{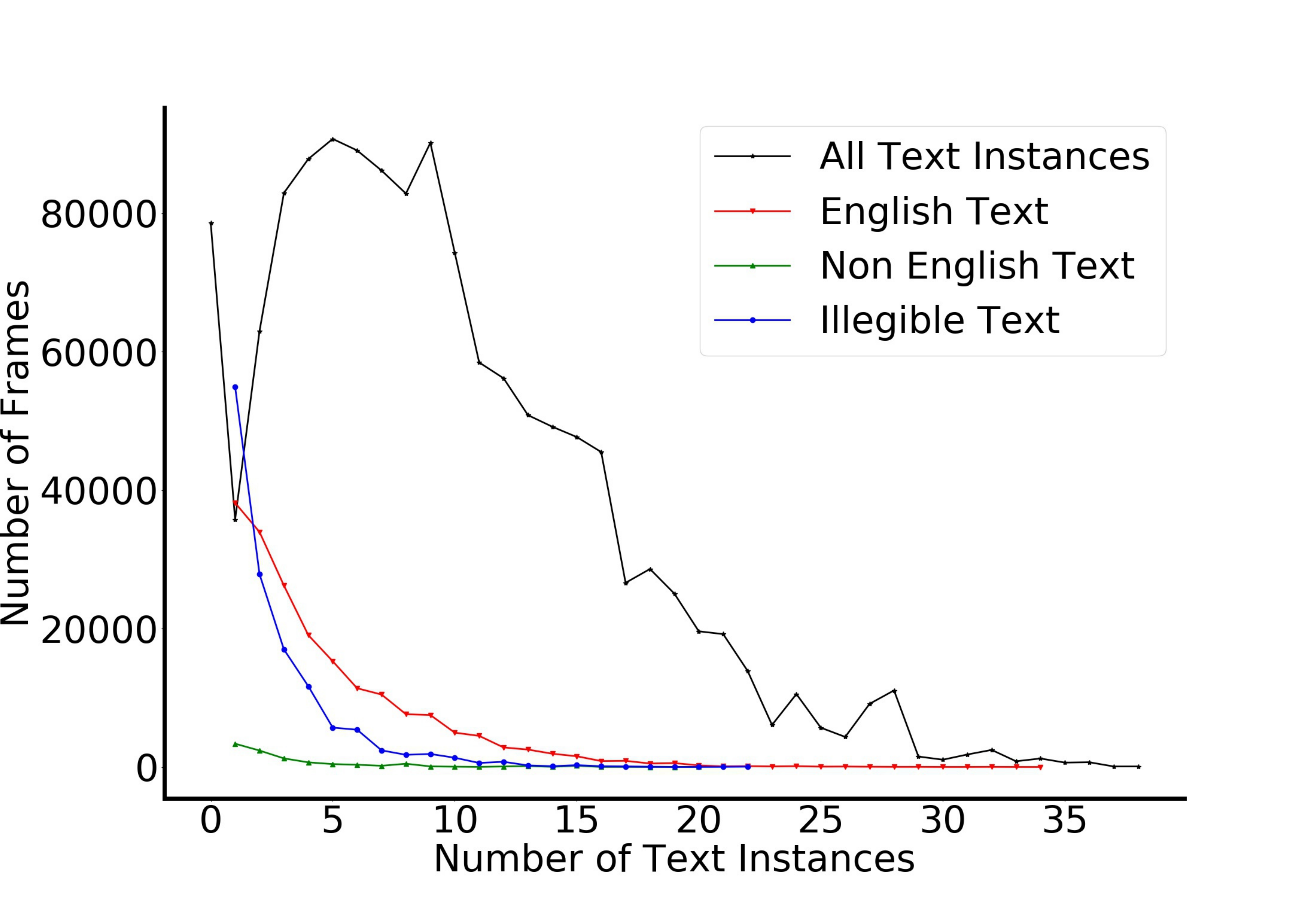}\centering \\ \small{(c) Number of frames with a particular number of text instances.}
\endminipage
\caption{\small Statistics of text instances in RoadText-1K.}
\label{fig_9}
\end{figure*}

\begin{figure*}[!htb]
\minipage{0.22\textwidth}
\hspace{-0.5mm}
  \includegraphics[width=\linewidth]{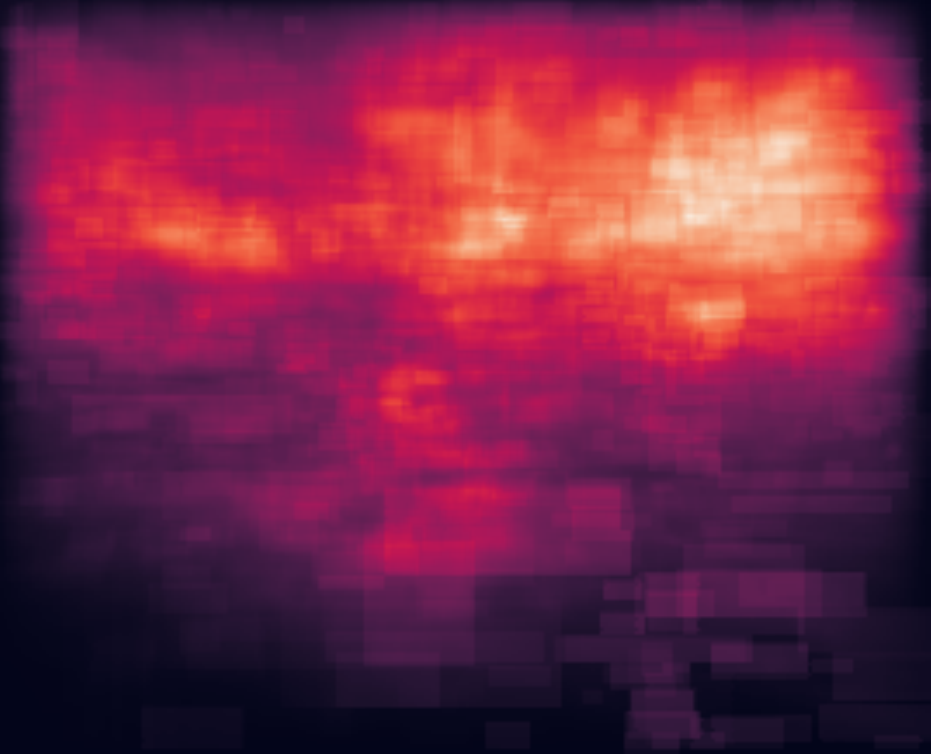} 
  \centering\\ \small{(a) RoadText-1K }
\endminipage\hfill
\minipage{0.22\textwidth}
  \includegraphics[width=\linewidth]{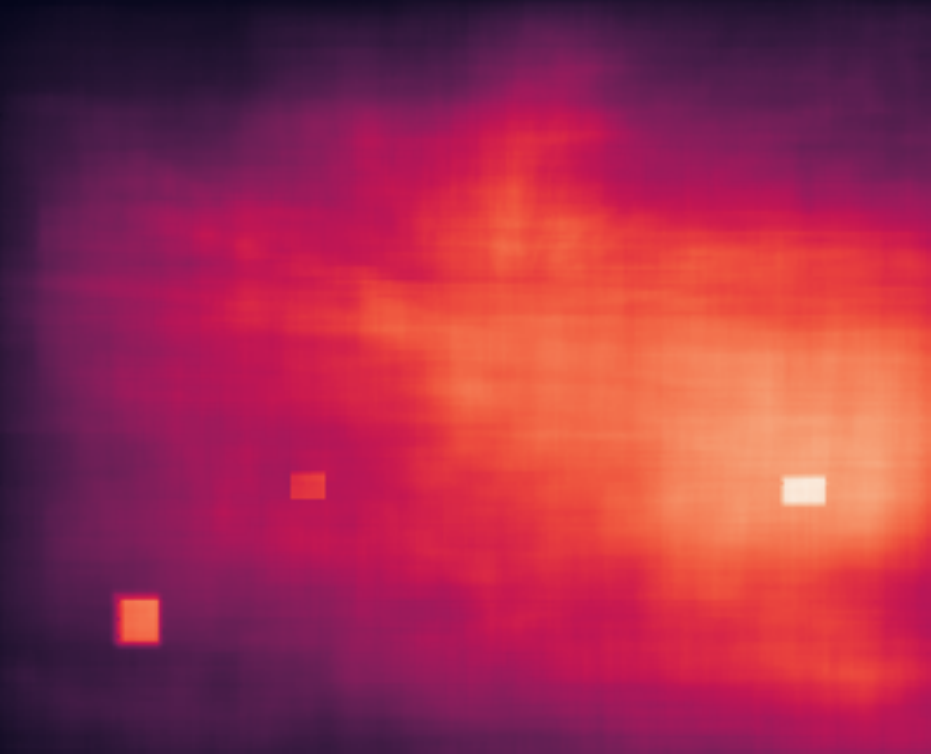}\centering\\\small{(b) Text in Videos}
\endminipage\hfill
\minipage{0.22\textwidth}%
  \includegraphics[width=\linewidth]{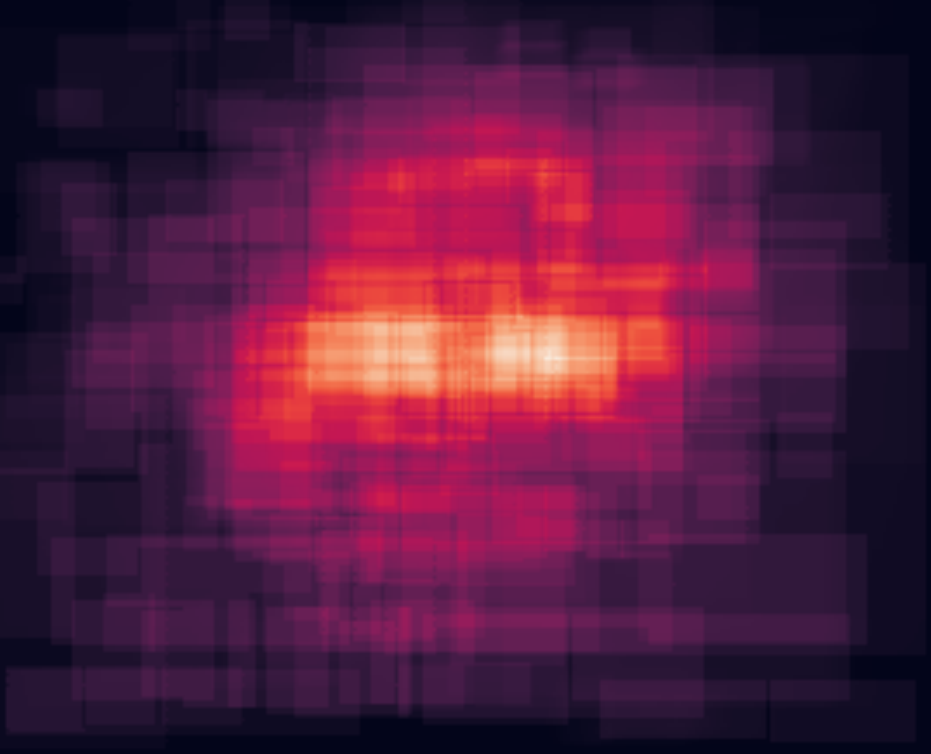}\centering\\\small{(c) Youtube Video Text}
\endminipage\hfill
\minipage{0.22\textwidth}%
  \includegraphics[width=\linewidth]{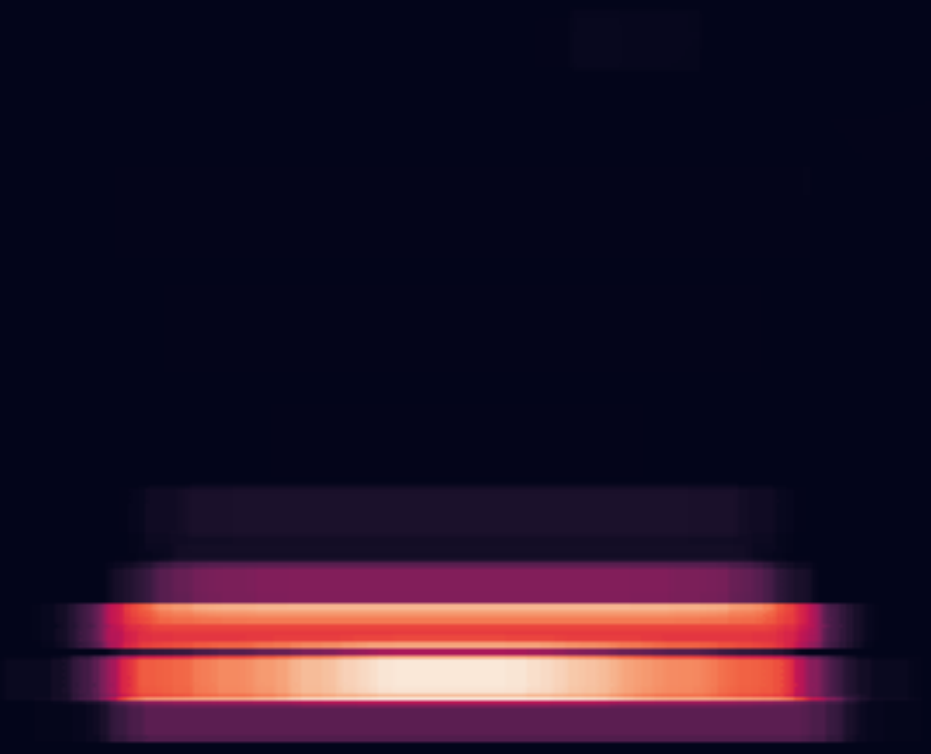}\centering\\\small{(d) USTB-VidText}
\endminipage
\setlength{\belowcaptionskip}{-1em}
\caption{\small Heatmaps depicting the spatial distribution of text instances for existing and proposed datasets.}
\label{fig_10}
\end{figure*}

Apart from video based datasets, Uber-Text~\cite{UberText} is large scale image dataset closest to the proposed RoadText-1K. This dataset is collected from images by the Bing Maps Streetside program largely aligning with our unconstrained driving videos from BDD100K. UberText contains 117969 street-level images with 571534 annotated text regions. The dataset also categorize text instances into 9 categories indicating whether the text is business name, street name etc. UberText aligns with the proposed dataset in terms of non text centered/focused frames, while the differences are in the bounding box shape. We provide rectangular bounding boxes while UberText provides free-hand polygon annotations for text instances.

\section{Experiments}
\subsection{Text Detection and Recognition in Images}
Various existing techniques for text detection and recognition are evaluated on the RoadText-1K, such as \textsc{ctpn}~\cite{EE:1}, \textsc{east}~\cite{DBLP:conf/cvpr/ZhouYWWZHL17} and \textsc{fots}~\cite{DBLP:conf/cvpr/LiuLYCQY18} for detection and \textsc{crnn}~\cite{EE:3} and \textsc{aster}~\cite{bshi2018aster}. \textsc{crnn} is quite popular for introducing the sequence to sequence model for text recognition and had significant performance improvement~\cite{EE:3} when proposed, while \textsc{aster} is currently state of the art in many popular datasets~~\cite{bshi2018aster}.  Since these techniques are developed for image, we generate still images from videos and the frame level annotations serve as its ground truth. 

The Connectionist Text Proposal Network (\textsc{ctpn}) model uses VGG16 backbone for feature extraction, followed by a Bi-directional \textsc{lstm} for output. Efficient and Accurate Scene Text Detector (\textsc{east}) is another state of the art text detector that uses fully convolutional layers for text prediction followed by Non-Maximum Suppression (\textsc{nms}). Fast Oriented Text Spotting (\textsc{fots}) an end-to-end model, shares convolutional layer's features for detection and recognition. The pretrained models of discussed text detection methods \textsc{ctpn}, \textsc{east} and \textsc{fots} have been trained on word images. While the current annotations are at line level making it not really a fair direct comparison. Hence we finetune the three models on the train set and evaluate on test set of RoadText-1K. Table~\ref{t3} below, shows the detection results.

\begin{table}[htb]
\caption{\small Frame level text detection results of existing models on the RoadText-1K.}
\label{t3}
\begin{center}
\scalebox{1.15}{
\begin{tabular}{|c|c|c|c|}
\hline
Method & Precision & Recall & F-score \\
\hline
CTPN & 0.44 & 0.41 & 0.42 \\
\hline
EAST  & 0.42 & 0.30 & 0.35 \\
\hline
FOTS & 0.45 & 0.36 & 0.40\\
\hline
\end{tabular}}
\end{center}
\end{table}

As mentioned earlier we use the existing \textsc{crnn}, \textsc{aster} methods to evaluate  recognition performance on frames. \textsc{crnn} is one of the most commonly used method for text recognition which uses a CNN to extract features followed by a Bi-directional \textsc{lstm} layer for modelling the sequence and finally \textsc{ctc} loss for training.
\textsc{aster} first uses a spatial transformer network~\cite{stn} to rectify the images followed by an encoder-decoder style recognition network to transcribe the rectified image.
These methods are evaluated on the new dataset and the results are shown in Table~\ref{t4}

\begin{table}[h]
\caption{\small Frame level text recognition accuracy of existing models on RoadText-1K, given  ground truth text line crops and tracks.}
\label{t4}
\begin{centering}
\scalebox{1.15}{
\begin{tabular}{|c|c|c|c|c|c|c|}
\hline
\multirow{2}{3em}{Method}  & \multicolumn{3}{c|}{Pretrained} & \multicolumn{3}{c|}{Fine tuned} \\
\cline{2-7} & All & AN & MV & All & AN & MV \\
\hline
CRNN & 29.0 & 44.6 & 60.1 & 36.3 & 50.9 & 65.2 \\
\hline
ASTER & 44.6 & 61.9 & 67.2 & 48.1 & 63.0 & 68.3\\
\hline
\end{tabular}}
\end{centering}
\end{table}

 In the Table~\ref{t4}, category ``\textsc{all}'' refers to text instances which are of English legible category including special characters. While ``\textsc{an}'' refers to alphanumeric which includes only alphanumeric characters, this is done to provide a fair comparison as pretrained models are in general trained only on alphanumeric data. ``\textsc{mv}'' refers to Majority Voting, where we consider the text transcription that has maximum occurrence across frames of the same text instance for evaluation.
 
From Tables~\ref{t3} and \ref{t4} we observe that frame level text detection and recognition results on RoadText-1K are not on par with the results these methods report on existing scene text datasets. For example \textsc{ctpn} which performs the best on RoadText-1K , reports an F-score of 0.88 on icdar-2013~\cite{DBLP:conf/icdar/KaratzasSUIBMMMAH13} benchmark compared to 0.42 on our dataset. Similarly, although the fine-tuned \textsc{aster} model yields $60\%+$ accuracy on our dataset, the same model reports $>90+$ on most benchmark datasets\cite{DBLP:conf/icdar/KaratzasSUIBMMMAH13,MishraBMVC12} for scene text recognition. The results suggest limitation of the current text detection and recognition methods on unconstrained, realistic driving videos, which are not text centric.

\subsection{Text Detection and Recognition in Videos}
We also evaluate video level methods which provide end to end results. The most common metrics used to evaluate are the \textsc{mot} metrics like \textsc{motp} (Multiple Object Tracking Precision), \textsc{mota} (Multiple Object Tracking Accuracy). These metrics are widely used in object tracking methods and the same has been used as a metric of evaluation in the \textsc{icdar} Robust Reading Challenge (2015) \cite{DBLP:conf/icdar/KaratzasSUIBMMMAH13} for End to End task in Text in Videos. We evaluate methods built for text tracking in video, object tracking and compare their performance.

The work \cite{DBLP:conf/icdar/WangJYZLFWL17} based on multi frame tracking provides a method to track text instances temporally based on attributes of the text objects in multiple frames. This tracking algorithm considers various factors like IOU, offset of matched frames, edit distance of text to link two text objects across frames to same track. This method has been evaluated with various combinations of text detection and recognition algorithms. The tracking results are presented in the Table~\ref{t5}.

\begin{table}[h]
\caption{\small Text Tracking performance on RoadText-1K using a method proposed in
\cite{DBLP:conf/icdar/WangJYZLFWL17}}
\label{t5}
\begin{centering}
\scalebox{0.95}{
\begin{tabular}{|c|c|c|c|c|c|c|}
\hline
\multirow{2}{3em}{Method}& \multicolumn{2}{c|}{MOTP} & \multicolumn{2}{c|}{MOTA} & \multicolumn{2}{c|}{ATA \%} \\
\cline{2-7} & CRNN & ASTER & CRNN & ASTER & CRNN & ASTER \\
\hline
CTPN & 17.06 & 7.4 & -29.79 & -11 & 0.56 & 0.57 \\
\hline
EAST & 11.48  & 11.5 & -111 & -111 & 0.36 & 0.31 \\
\hline
FOTS & 10.75 & 11 & -206 & -206 & 0.13 & 0.10\\
\hline
\end{tabular}}
\end{centering}
\end{table}
In the recent past object tracking community has been using new evaluation metrics for tracking presented in the CVPR19: Tracking and Detection Challenge \cite{DBLP:journals/corr/abs-1906-04567}. These new metrics include \textit{IDF1} (ID F1-score) ratio of correctly identified detections over the average number of ground-truth and computed detections, \textit{\textsc{mt}} (Mostly Tracked) Number of objects tracked for at least 80 percent of lifespan, \textit{\textsc{ML}} (Mostly Lost) Number of objects tracked less than 20 percent of lifespan, \textit{Ids} Number of ID switches across tracks, \textit{\textsc{fm}} Total number of times a trajectory is fragmented along with \textit{\textsc{mota}} Multiple Object Tracking Accuracy, \textit{\textsc{fp}}  False Positives, \textit{\textsc{fn}} False Negatives. We also evaluate the metrics provided in the challenge using one of the popular object tracking methods \textsc{sort} \cite{DBLP:conf/icip/BewleyGORU16}.

We use detections from various methods and compare the results in Table~\ref{t6}.
\begin{table}[h]
\centering
\caption{\small Evaluation of Text Tracking by SORT \cite{DBLP:conf/icip/BewleyGORU16} using new MOT evaluation metrics proposed in CVPR 2019 Tracking and Detection challenge \cite{DBLP:journals/corr/abs-1906-04567}}
\label{t6}
\begin{centering}
\scalebox{1.0}{
\begin{tabular}{|c|c|c|c|}
\hline
Metric & CTPN & EAST & FOTS \\
\hline
MOTA & -65.3 & -122.7 &-195.0\\
\hline
IDF1 (\%) & 9.7 & 12.2 & 12.15\\
\hline
MT & 1.2 & 1.23 &1.82\\
\hline

ML & 39.03 & 23.3 &27.8\\
\hline
FP & 470 & 470.7 &492.6\\
\hline
FN & 628 & 601.6 & 654.6 \\
\hline
IDs & 28.3 & 22.12 & 22.89\\
\hline
FM & 23.7 & 19.26  &19.36\\

\hline
\end{tabular}}
\end{centering}
\end{table}

\section{Conclusion}
We motivate the problem of understanding text in driving videos and introduce a new large-scale dataset for the same. We find that existing datasets for text spotting in images and videos are usually curated with text in mind and hence do not fare well on a realistic setting, like videos captured from a moving car. We believe that RoadText-1K will encourage research both on improving text detection and recognition in videos as well as enabling and equipping complex tasks like self-driving and driver assistance which can hugely benefit from reasoning about text on the road.

\bibliographystyle{IEEEtran}
\bibliography{IEEEabrv,IEEEexample}
\addtolength{\textheight}{-12cm}

\end{document}